\documentclass[10pt,conference,a4paper]{IEEEtran}

\usepackage{times}
\usepackage{tikz}
\usepackage{balance}
\usepackage{multirow}
\usepackage{multicol}

\usepackage[utf8]{inputenc}
\usepackage[english]{babel}
\usepackage{booktabs}
\usepackage[T1]{fontenc}
\usepackage{epsfig}
\usepackage{graphicx}
\usepackage{flushend}
\usepackage{amssymb,amsmath}
\usepackage{fixltx2e}
\newcommand{\ra}[1]{\renewcommand{\arraystretch}{#1}}

\begin{document}
%

\title{Analyzing features learned for Offline Signature Verification using Deep CNNs}

\author{

\IEEEauthorblockN{Luiz G. Hafemann, Robert Sabourin}
\IEEEauthorblockA{Lab. d'imagerie, de vision et d'intelligence artificielle\\
\'Ecole de technologie sup\'erieure\\
Universit\'e du Qu\'ebec, Montreal, Canada\\
lghafemann@livia.etsmtl.ca, robert.sabourin@etsmtl.ca}

\and

\IEEEauthorblockN{Luiz S. Oliveira}
\IEEEauthorblockA{Department of Informatics\\
Federal University of Parana\\
Curitiba, PR, Brazil\\
lesoliveira@inf.ufpr.br}

\and

}

\usetikzlibrary{calc}
\maketitle
\begin{tikzpicture}[remember picture, overlay]
\node at ($(current page.north) + (0,-0.5in)$) {\large Accepted as a conference paper for ICPR 2016};
\end{tikzpicture}

\begin{abstract}
Research on Offline Handwritten Signature Verification explored a large variety of handcrafted feature extractors, ranging from graphology, texture descriptors to interest points. In spite of advancements in the last decades, performance of such systems is still far from optimal when we test the systems against skilled forgeries - signature forgeries that target a particular individual.  In previous research, we proposed a formulation of the problem to learn features from data (signature images) in a Writer-Independent format, using Deep Convolutional Neural Networks (CNNs), seeking to improve performance on the task. 
In this research, we push further the performance of such method, exploring a range of architectures, and obtaining a large improvement in state-of-the-art performance on the GPDS dataset, the largest publicly available dataset on the task. In the GPDS-160 dataset, we obtained an Equal Error Rate of 2.74\%, compared to 6.97\% in the best result published in literature (that used a combination of multiple classifiers). We also present a visual analysis of the feature space learned by the model, and an analysis of the errors made by the classifier. Our analysis shows that the model is very effective in separating signatures that have a different global appearance, while being particularly vulnerable to forgeries that very closely resemble genuine signatures, even if their line quality is bad, which is the case of slowly-traced forgeries.

\end{abstract}


\IEEEpeerreviewmaketitle

\section{Introduction}

The handwritten signature is a widely accepted means of authentication in government, legal, and commercial transactions.  Signature verification systems aim to confirm the identity of a person based on their signature \cite{jain_introduction_2004}, that is, they classify signature samples as ``genuine'' (created by the claimed individual) or ``forgery'' (created by an impostor). 
In offline (static) signature verification, the signatures are acquired after the signature writing process is completed, by scanning a document containing the signature. This is in contrast with online (dynamic) signature verification, where the signature is captured directly on a device (such as a pen tablet), and therefore the dynamic information of the signature is available, such as the velocity of the pen movements. The lack of dynamic information in the offline case makes it a challenging problem, and much of the effort in this field has been devoted to obtaining a good feature representation for signatures \cite{hafemann_offline_2015}.

There are two main approaches for the problem in the literature: in a Writer-Dependent approach, for each user, a training set of genuine signatures of the user (and, often, genuine signatures from other users as negative samples) is used to train a binary classifier. 
In a Writer-Independent approach, a single global classifier is trained using a dissimilarity approach, by using a training set consisted of positive samples (a difference vector between two genuine signatures from the same author), and negative samples  (a difference vector between a genuine signature and a forgery). At test time, a distance vector is calculated between the query signature and one or more reference signatures (known to be genuine signatures of the claimed individual), and then classified using the Writer-Independent classifier. For a comprehensive review on the problem, refer to  \cite{impedovo_automatic_2008}, and for a recent review, see \cite{hafemann_offline_2015}.

Recent work on the area explore a variety of different feature descriptors: Extended Shadow Code (ESC) and Directional-Probabilistic Density Function (DPDF) \cite{rivard_multi-feature_2013}, \cite{eskander_hybrid_2013}; Local Binary Patterns (LBP), Gray-Level Co-occurrence Matrix (GLCM) and Histogram of Oriented Gradients (HOG) \cite{yilmaz_offline_2015}, \cite{hu_offline_2013}; Curvelet transform \cite{guerbai_effective_2015}, among others. Instead of relying on hand-engineered feature extractors, we investigate feature learning algorithms applied to this task. In previous research \cite{hafemann_ijcnn_2016}, we have shown that we can learn useful features for offline signature verification, by learning Writer-Independent features from a development dataset (a set of users not enrolled in the system). Using this formulation, we obtained results close to the state-of-the-art in the GPDS dataset, when compared against models that rely on a single feature extractor technique. In the same vein, Ribeiro et al. \cite{ribeiro_deep_2011} used Restricted Boltzmann Machines (RBMs) for learning features from signature images. However, the authors considered only a small set of users (10), and did not use the features to actually classify signatures, only reporting a visual representation of the learned weights. Khalajzadeh \cite{khalajzadeh_persian_2012} used Convolutional Neural Networks for Persian signature verification, but did not considered skilled forgeries.

In this paper, we further analyze the method introduced in \cite{hafemann_ijcnn_2016}, investigating the impact of the depth (number of layers) of the Deep Convolutional Neural Network (CNN), and the size of the embedding layer on learning good representations for signatures, as measured by the classification performance on a different set of users. Using better training techniques (in particular, using Batch Normalization \cite{ioffe2015batch} for training the network), we can improve performance significantly, achieving a state-of-the-art performance of 2.74\% Equal Error Rate (EER) on the GPDS-160 dataset, which surpasses all results in the literature, even comparing to results where model ensembles are used.
We also perform an analysis on the errors committed by the model, and visualize the quality of the learned features with a 2D projection of the embedding space. Visual analysis suggests that the learned features capture the overall aspect of the signature, which is sufficient to perform well in separating genuine signatures from skilled forgeries in some cases: for users that have complex signatures, or that maintain very stable signatures (i.e. different genuine samples are very similar). We also notice that the learned features are particularly vulnerable to slowly-traced forgeries, where the overall signature shape is similar, but the line quality is poor.

\section{Methodology}

The central idea is to learn a feature representation (a function $\phi(.)$) for offline signature verification in a Writer-Independent format, use this function to extract features from signatures $\textbf{X}$ of the users enrolled in the system ($\phi(\textbf{X})$), and use the resulting feature vectors to train a binary classifier for each user. The rationale for learning the feature representation in a Writer-Independent format is two-fold: 1) learning feature representations directly for each user is impractical, given the low number of samples available for training (around 5-10 signatures); and 2) having a fixed representation useful for any user makes it straightforward to add new users to the system, by simply using the model to extract features for the new user's genuine signatures, and training a Writer-Dependent classifier.

In order to learn features in a Writer-Independent format, we consider two separate sets of signatures: a development set $\mathcal{D}$, that contains signatures from users not enrolled in the system, and an exploitation set $\mathcal{E}$ that contains a disjoint set of users. The signatures from set $\mathcal{D}$ are only used for learning a feature representation for signatures, with the hypothesis that the features learned for this set of users will be relevant to discriminate signatures from other users. On the other hand, the set $\mathcal{E}$ represents the users enrolled to the system. We use the model trained on the set $\mathcal{D}$ to ``extract features'' for the signatures of these users, and train a binary classifier for each user. It is worth noting that we do not use skilled forgeries during training, since it is not practical to require forgeries for each new user enrolled in the system.

Overall, the method consists in the following steps:
\begin{itemize}
\item Training a deep neural network on a Development set
\item Using this network to obtain a new representation for signatures on $\mathcal{E}$ (i.e. obtain $\phi(X)$ for all signatures $X$)
\item Training Writer-Dependent classifiers in the exploitation set, using the learned representation
\end{itemize}

Details of these steps are presented below.

\subsection{Convolutional Neural Network training}

As in \cite{hafemann_ijcnn_2016}, we learn a function $\phi(.)$ by training a Deep Convolutional Neural Network on a Development set, by learning to discriminate between different users. That is, we model the network to output $M$ units, that estimate $P(y|X)$ where $y$ is one of the $M$ users in the set $\mathcal{D}$, and $X$ is a signature image. In this work we investigate different architectures for learning feature representations. In particular, we evaluate the impact of depth, and the impact of the size of the embedding layer (the layer from which we obtain the representation of the signature). Multiple studies suggest that depth is important to address complex learning problems, with both theoretical arguments \cite{bengio_learning_2009}, and as shown empirically (e.g. \cite{simonyan2014very}). The size of the embedding layer is an important factor for practical considerations, since this is the size of the feature vectors that will be used for training classifiers for each new user of the system, as well as performing the final classification.

In our experiments, we had difficulty to train deep networks, with more than 4 convolutional layers and 2 fully-connected layers, even using good initialization strategies such as recommended in \cite{glorot_understanding_2010}. Surprisingly, the issue was not that the network overfit the training set, but rather both the training and validation losses remained high, suggesting issues in the optimization problem. To address this issue, we used Batch Normalization \cite{ioffe2015batch}. This technique consists in normalizing the outputs of a layer, for each individual neuron, so that the values have zero mean and unit variance (in a mini-batch of training data), and showed to be fundamental in our experiments in obtaining good performance. Details of this technique can be found in \cite{ioffe2015batch}. In the experiments for this paper, we only report the results using Batch Normalization, since for most of the proposed architectures, we could not train the network without this technique (performance on a validation set remained the same as random chance).

\begin{table}
\ra{1.3}
\centering
\caption{CNN architectures evaluated in this paper}
\label{tbl:architectures}
\begin{tabular}{|c|c|c|c|}
\hline 
AlexNet\_\textbf{N}\textsubscript{reduced}&AlexNet\_\textbf{N}& 
VGG\_\textbf{N}\textsubscript{reduced}&VGG\_\textbf{N} \\ \hline
conv11-96-s4-p0 & conv11-96-s4-p0 & conv3-64 & conv3-64 \\
& & conv3-64 & conv3-64 \\ \hline
pool3-s2-p0 & pool3-s2-p0 & pool3 & pool3 \\ \hline
conv5-256-p2 & conv5-256-p2 & conv3-128 & conv3-128\\
& & conv3-128 & conv3-128 \\ \hline
pool3-s2-p0 & pool3-s2-p0  & pool4 & pool3 \\ \hline
conv3-384 & conv3-384 & conv3-256 & conv3-256 \\
conv3-256 & conv3-384  & conv3-256 & conv3-256 \\
     & conv3-256  & conv3-256 & conv3-256 \\
 	 & 		& conv3-256 & conv3-256 \\ \hline
pool3-s2-p0 & pool3-s2-p0 & pool4 & pool3 \\ \hline
& & & conv3-256  \\
& & & conv3-256  \\
& & & conv3-256  \\
& & & conv3-256  \\ \cline{4-4}
& & & pool2 \\ \cline{4-4}
& & & conv3-256  \\
& & & conv3-256 \\
& & & conv3-256  \\
& & & conv3-256  \\ \cline{4-4}
& & & pool2 \\ \cline{4-4}
& \textbf{FC1-N} & & \textbf{FC1-N} \\ \cline{2-2} \cline{4-4}
\textbf{FC1-N} &\textbf{FC2-N} &\textbf{FC1-N} &\textbf{FC2-N} \\ \hline
\multicolumn{4}{|c|}{FC-531 + softmax} \\ \hline
\end{tabular}
\end{table}

Table \ref{tbl:architectures} shows the architectures we evaluated in this research. The models are based on AlexNet \cite{krizhevsky_imagenet_2012} and VGG \cite{simonyan2014very}, which are two architectures that perform remarkably well in Computer Vision problems. Each row specifies a layer of the network. For convolutional layers, we include first the filter-size, followed by number of convolutional filters. For pooling layers, we include the size of the pooling region. Unless otherwise indicated, we use stride 1 and padding 1. For instance, conv11-96-s4-p0 refers to a convolutional layer with 96 filters of size 11x11, with stride 4 and no padding. For the Fully Connected (FC) layers, we just indicate the name of the layer (that we refer later in the text), and the number of output units. For instance, FC1-N refers to a Fully connected layer named ``FC1'', with $N$ output units. The network AlexNet\_N\textsubscript{reduced} contains 9 layers: 6 layers with learnable parameters (convolutions and fully connected layers), plus 3 pooling layers, where the VGG\_N network contains 24 layers: 19 layers with learnable parameters plus 5 pooling layers. We used Batch Normalization for each learnable layer in the network, before applying the non-linearity (ReLU, except for the last layer, which uses softmax). We considered 4 types of networks: AlexNet: a network similar to the network proposed in \cite{krizhevsky_imagenet_2012}, but without Local Response Normalization and Dropout (which we found unnecessary when using Batch Normalization); AlexNet\textsubscript{reduced}: a similar network but with reduced number of layers; VGG: An architecture similar to \cite{simonyan2014very}; VGG\textsubscript{reduced}: a similar network but with reduced number of layers. For each network, we consider a different size $N$ for the embedding layer, $N \in \{512, 1024, 2048, 4096\}$. This is the size of the feature vectors that will be used for training the Writer-Dependent classifiers, and therefore they impact the cost of training classifiers for new users that are enrolled to the system, as well as the classification cost. Considering the different architectures, and size of the embedding layer, we tested in total 16 architectures. 

The images were pre-processed by: centralizing the signatures in a 840x1360 image according to their center of mass; removed the background using OTSU's algorithm; inverted the images so that the background pixels were zero valued; and finally resizing them to 170x242 pixels. More details on these preprocessing steps can be found in \cite{hafemann_ijcnn_2016}.


\subsection{Training the Writer-Dependent classifiers}

After training a Deep CNN on the development set $\mathcal{D}$, we use the network to extract feature representations for signatures from another set of users ($\mathcal{E}$) and train Writer-Dependent classifiers. To do so, we resize the images to 170x242 pixels, perform feedforward propagation until a fully-connect layer, and used the activations at that layer as the feature vector for the image. For the architectures marked as ``reduced'' (see table \ref{tbl:architectures}), that contained only two fully-connected layers, we consider only the last layer before softmax as the embedding layer (FC1). For the architectures with three fully-connected layers, we consider both layers FC1 and FC2, that is, the last two fully-connected layers before the softmax layer.


For each user in the set $\mathcal{E}$, we build a training set consisted of $r$ genuine signatures from the user as positive samples, and 14 genuine signatures from users in $\mathcal{D}$ as negative samples. We train a Support Vector Machine (SVM) classifier on this dataset, considering both linear SVM, and SVM with an RBF kernel. Similarly to \cite{eskander_hybrid_2013}, we use different weights for the positive and negative class to account for the imbalance of having many more negative samples than positive. We do so by duplicating the positive samples, so that the dataset is roughly balanced. For testing, we use  10 genuine signatures from the user (not used for training), 10 random forgeries (signatures from other users in $\mathcal{E}$), and the 30 skilled forgeries made targeting the user's signature.

\section{Experimental Protocol}

We conducted experiments in the GPDS-960 dataset \cite{vargas_off-line_2007}, which is the largest publicly available dataset for offline signature verification. The dataset consists of 881 users, with 24 genuine samples per user, and 30 skilled forgeries.

We divide the dataset into an exploitation set $\mathcal{E}$ consisting of the first 300 users, and a development set consisting of the remaining 581 users. Since in this work we evaluate many different architectures, we split the Development set into a training and validation set, using a disjoint set of users for each set. We use a subset of 531 users for training the CNN models, and a subset of 50 users for validation. We use the same protocol above to train Writer-Dependent classifiers for the validation set, to obtain an estimate of how well the features learned by the models generalize to other users. Finally, we use the best models to train WD classifiers for users in the exploitation set $\mathcal{E}$. We performed 10 runs, each time training a network with a different initialization, and using a random split of the genuine signatures used for training/testing, when training the Writer-Dependent classifiers. We report the mean and standard deviation of the errors across the 10 runs. To compare with previous work, we use as exploitation set the first 160 users (to compare with work that used the dataset GPDS-160), and the first 300 users (to compare with work that used GPDS-300). To assess the impact of the number of genuine signatures for each user in training, we ran experiments with 5 and 14 genuine signatures for training ($r \in \{5,14\}$). We evaluated the results in terms of False Rejection Rate (FRR - the fraction of genuine signatures misclassified as forgery), False Acceptance Rate (FAR - the fraction of forgeries misclassified as genuine signatures). For completeness, we show the FAR results for both random and skilled forgeries. We also report Equal Error Rates (EER), which is the error when FAR = FRR. In this case, we use only genuine signatures and skilled forgeries, to enable comparison with other work in the literature. We considered two forms of calculating the EER: EER\textsubscript{user-thresholds}: using user-specific decision thresholds; and EER\textsubscript{global threshold}: using a global decision threshold. Lastly, we also report the Mean Area Under the Curve (Mean AUC), by averaging the AUCs of Receiving Operating Curves (ROC curves) calculated for each user. For calculating FAR and FRR in the exploitation set, we used a decision threshold selected from the validation set (the threshold that achieved EER using a global decision threshold).

\section{Results}

\begin{figure}
\centering
\includegraphics[width=\columnwidth]{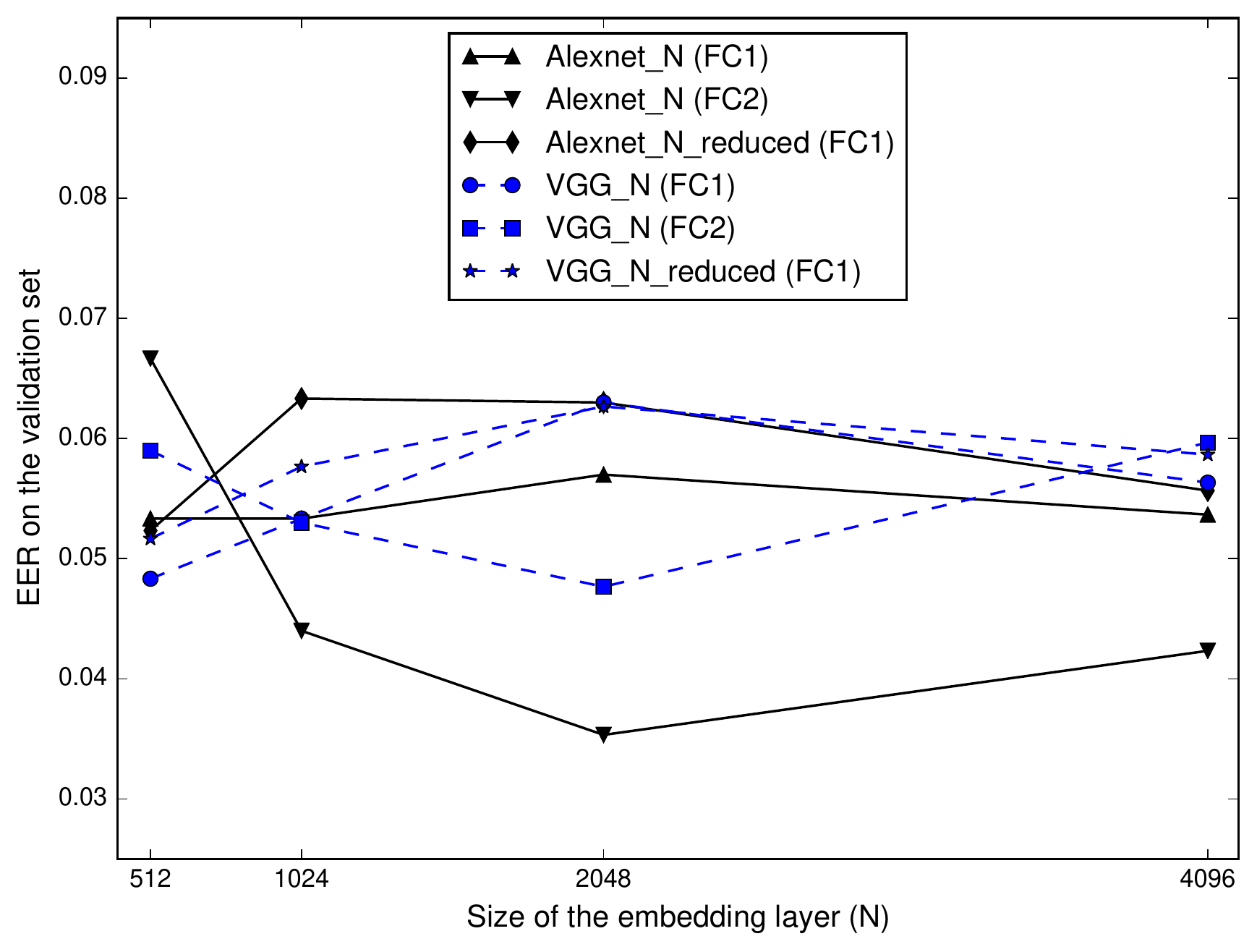}
\caption{Equal Error Rates on the validation set, for WD models trained with a \textbf{Linear SVM}, using the representation space learned by different architectures (at the layer indicated in parenthesis), and different representation sizes (N)}
\label{fig:varying_N_linear}
\end{figure}

\begin{figure}
\centering
\includegraphics[width=\columnwidth]{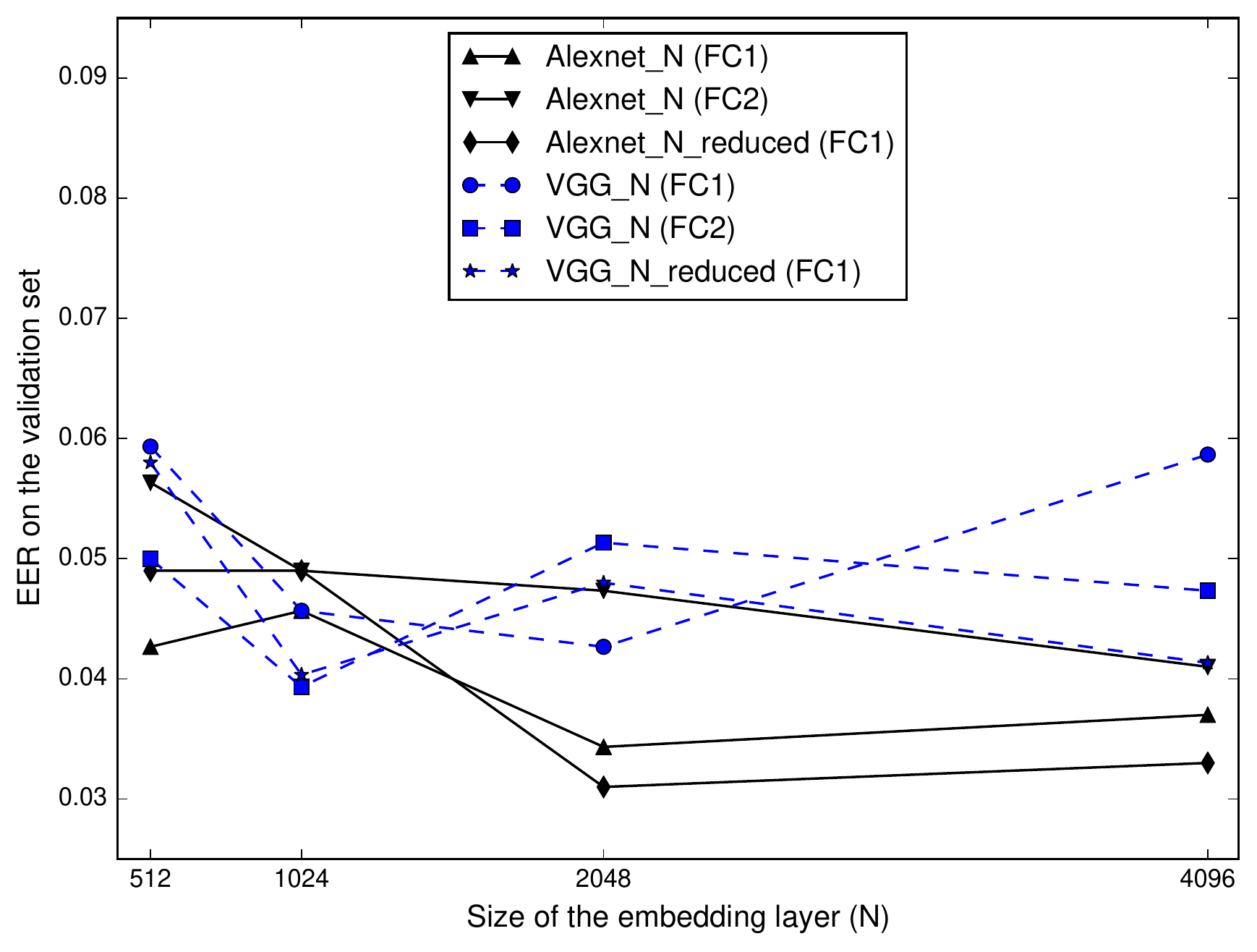}
\caption{Equal Error Rates on the validation set, for WD models trained with an SVM with \textbf{RBF kernel}, using the representation space learned by different architectures (at the layer indicated in parenthesis), and different representation sizes (N)}
\label{fig:varying_N_rbf}
\end{figure}

We first consider the results of the experiments in the validation set, varying the depth of the networks, and the size of the embedding layer. In these experiments, we trained the CNN architectures defined in table \ref{tbl:architectures}, used them to extract features for the users in the validation set, and trained Writer-Dependent classifiers for these users, using 14 reference signatures. We then analyzed the impact in classification performance of the different architectures/sizes of the embedding layer, measuring the average Equal Error Rate of the classifiers. Figures \ref{fig:varying_N_linear} and \ref{fig:varying_N_rbf} show the classification results on the validation set using Linear SVMs and SVMs with an RBF kernel, respectively. The first thing we notice is that, contrary to empirical results in object recognition (ImageNet), we did not observe improved performance with very deep architectures. The best performing models were the AlexNet architecture (with 8 trainable layers) and the AlexNet\textsubscript{reduced} (with 6 trainable layers) when using the features to training linear SVMs and SVMs with RBF kernel, respectively. We also notice that the performance with a linear classifier is already quite good, demonstrating that the feature representations learned in a writer-independent way seems to generalize well to new users.

\begin{figure}
\centering
\includegraphics[width=\columnwidth]{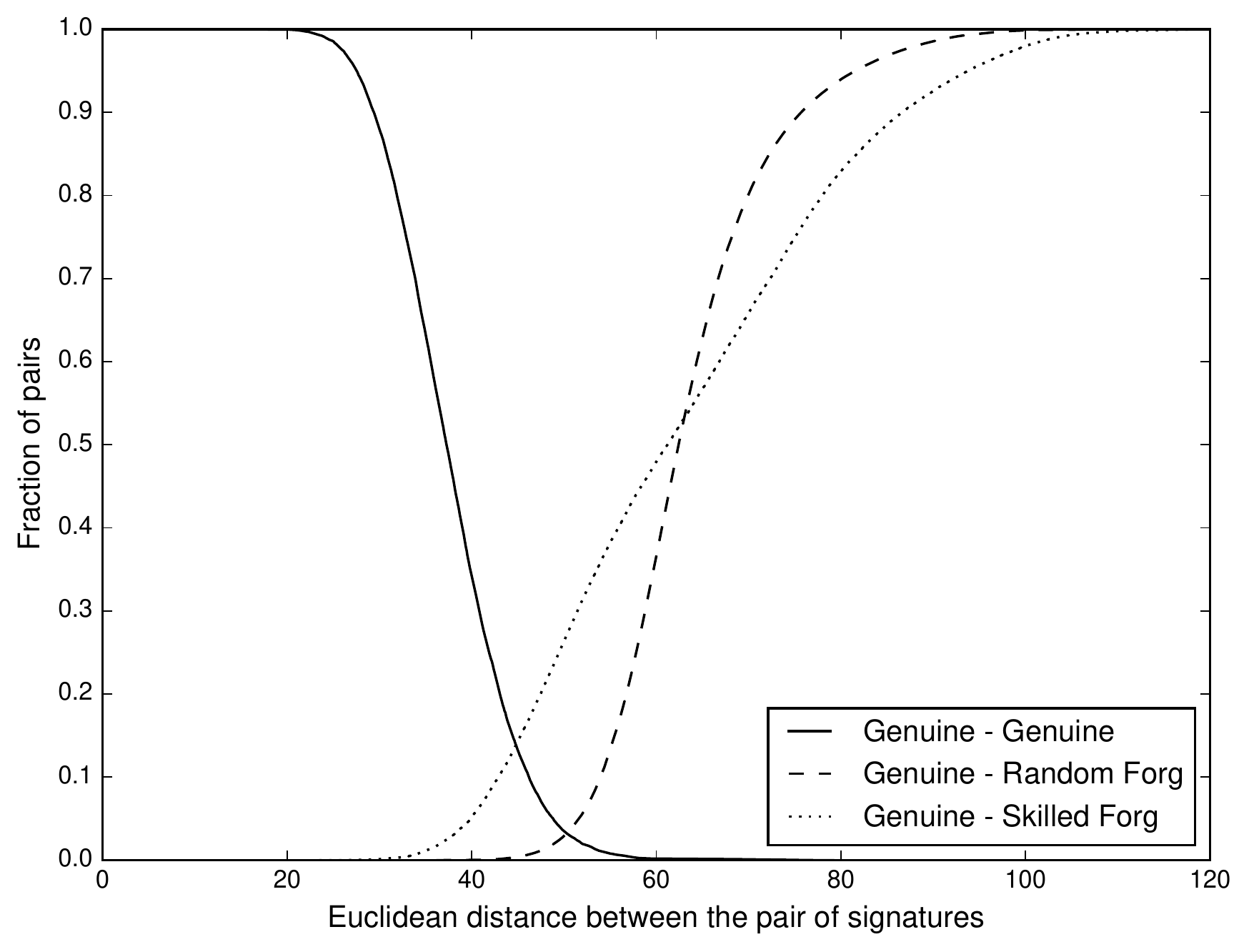}
\caption{Cumulative frequencies of pairs of signatures within a given distance. For the genuine-forgery pairs, the y axis shows the fraction of pairs that are \textit{closer} than a given euclidean distance. For genuine-genuine pairs  the y axis shows the fraction of pairs that are \textit{further} than a given distance, to show the overlap with the genuine-forgery pairs.}
\label{fig:distances}
\end{figure}

\begin{figure}
\centering
\includegraphics[scale=0.35]{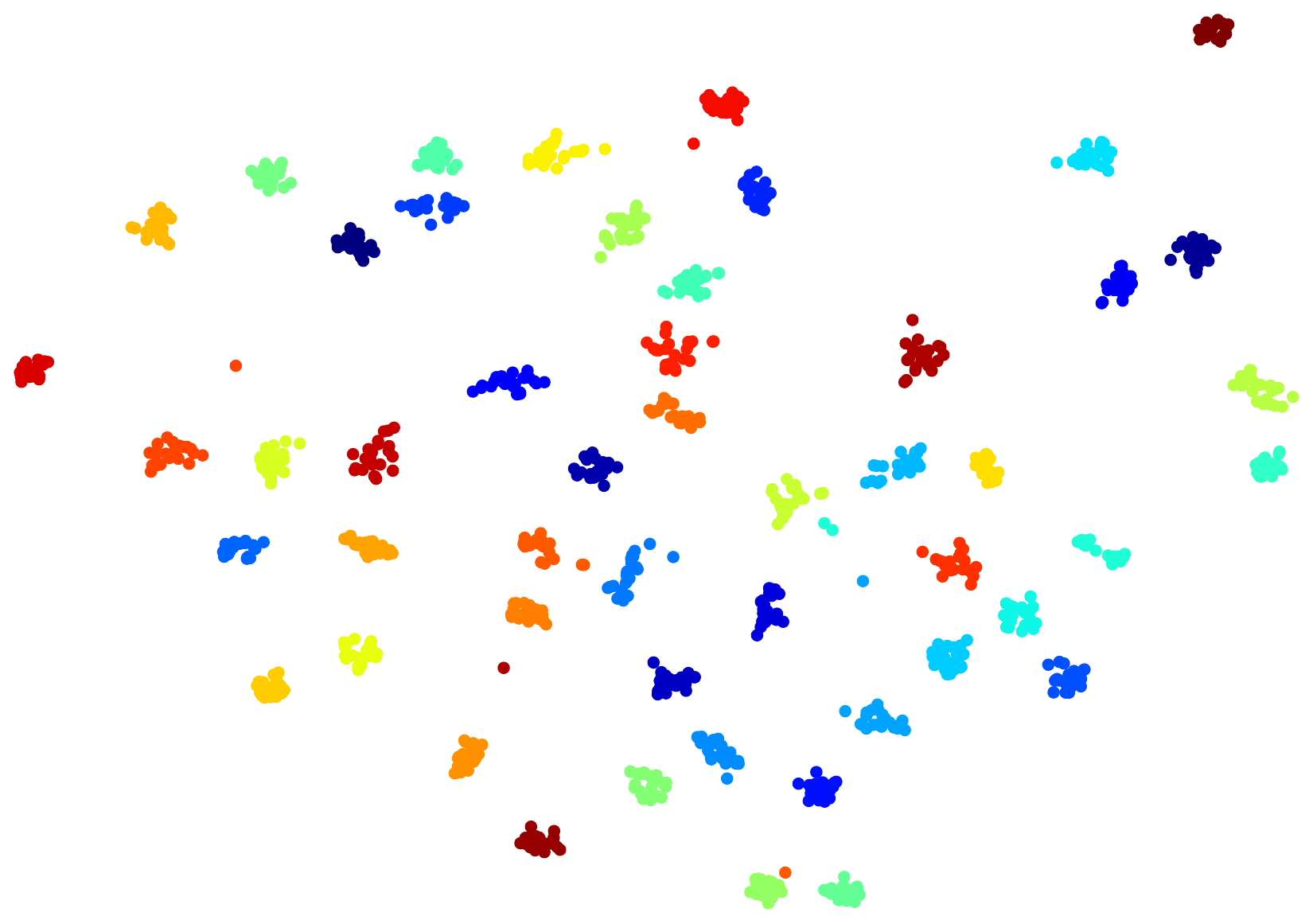}
\caption{2D t-SNE projection of the embedding space $\phi(.)$ for the genuines signatures of the 50 authors in the validation set. Each point refers to one signature, and different users are shown in different colors}
\label{fig:tsne_genuine}
\end{figure}

\begin{figure}
\centering
\includegraphics[width=\columnwidth]{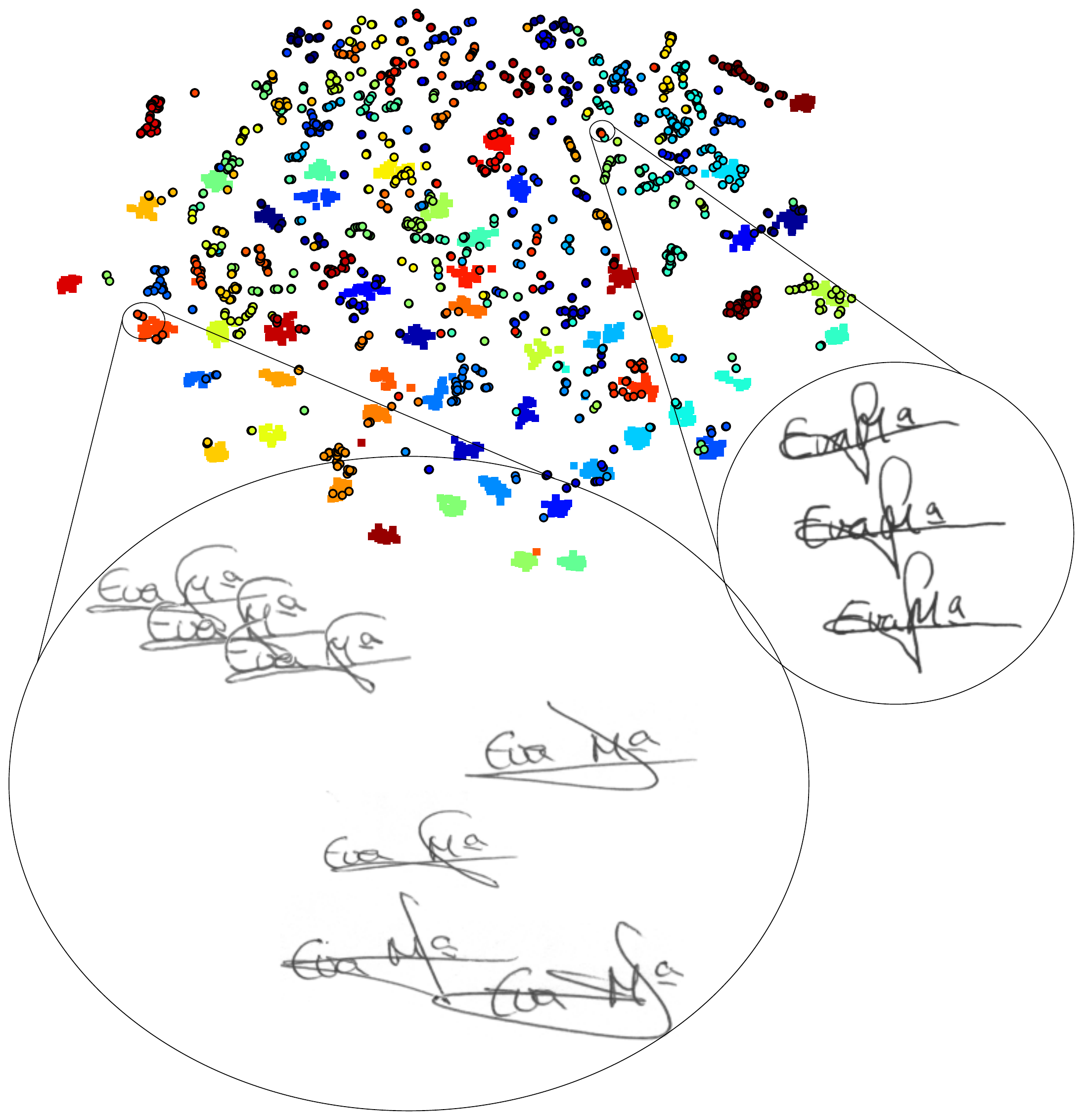}
\caption{2D t-SNE projection of the embedding space $\phi(.)$ for the genuine signatures and skilled forgeries of the 50 authors in the validation set. Points that refer to skilled forgeries are shown with a black border. The zoomed area on the left shows a region where three skilled forgeries (on the top-left) are close to four genuine signatures for user 394 in the GPDS dataset. The zoomed area on the right shows three skilled forgeries for the same user that are far from the genuine signatures in the embedding space}
\label{fig:tsne_genuine_fogery}
\end{figure}

Another aspect that we investigated is how the signatures from the users in the validation set are represented in the embedding layer of the CNN. For this analysis, we took one of the models (AlexNet\_2048, at layer FC2), obtained the representations for signatures of all 50 users in the validation set, and analyzed this representation. We started by analyzing how well separated are the signatures from each user in this feature space, as well as compared to skilled forgeries targeted to the users. To do so, we measured the euclidean distance between genuine signatures of the same user, genuine signatures from different users (i.e. a genuine signature vs. a random forgery), and the distance between pairs of a genuine signature and a skilled forgery made for this user. Figure \ref{fig:distances} plots the cumulative frequency of pairs of signatures within a given distance, that is, the fraction of pairs that are closer than a given euclidean distance in the embedding space. For the genuine-genuine pairs, we show the inverse of the cumulative frequency (i.e. the fraction of pairs further than a given distance), to better show the overlap between the curves. For a particular distance, the y axis is the expected False Rejection Rate (on the genuine-genuine pairs) and False Acceptance Rate (on the genuine-forgery pairs) if we used a distance-based classifier, with a single genuine signature as reference. We can see that in the space projected by the CNN, the genuine signatures from different users very well separated: there is a very small overlap between the curves of genuine-genuine pairs and genuine-random forgery pairs. A similar behavior can be seem with the genuine-skilled forgeries pairs, although the curves overlap more in this case. 

To further analyze how the signatures are dispersed in this representation space, we used the t-SNE algorithm \cite{van2008visualizing} to project the samples from $\mathbb{R}^{2048}$ to $\mathbb{R}^{2}$. This allows us to inspect the local structure present in this higher dimensionality representation. Figure \ref{fig:tsne_genuine} shows the result of this projection of the genuine signatures of the 50 users in the validation set. Each point is a signature, and each color represent a different user. We can see that the signatures from different users are very well separated in this representation, even though no samples from these authors were used to train the CNN, which suggests that the learned representation is not specific to the writers in the training set, but generalize to other users. Figure \ref{fig:tsne_genuine_fogery} shows the projection of both the genuine signatures and skilled forgeries for these 50 users. We can see that the skilled forgeries are more scattered around, and for several users, the skilled forgeries are close to the representation of the genuine signatures.

\begin{table*}
\centering
\caption{Detailed performance of the WD classifiers on the GPDS dataset  (Errors and Standard deviations in \%)} 
\label{table:results}
\resizebox{\textwidth}{!}{%
\begin{tabular}{llllrrrrrr}
\toprule
Dataset&  \begin{tabular}[x]{@{}c@{}}\#samples\\per user\end{tabular}  & Model & Classifier &      FRR &  FAR\textsubscript{random} &  FAR\textsubscript{skilled} &    EER\textsubscript{global threshold} &  EER\textsubscript{user-thresholds} & Mean AUC \\
\midrule
GPDS-160 & 5 & AlexNet\_2048 & Linear SVM  & 32.12 (+-1.21) &        0.01 (+-0.02) &         0.89 (+-0.13) &                 8.24 (+-0.54) &  4.25 (+-0.37) &   0.9845 (+-0.0018) \\

&    & AlexNet\_2048\textsubscript{reduced}  & SVM (RBF)  & 26.56 (+-1.01) &        0.00 (+-0.00) &         1.03 (+-0.14)&                 6.99 (+-0.26) & 3.83 (+-0.33)  &   0.9861 (+-0.0015) \\

& 14 & AlexNet\_2048  & Linear SVM & 10.41  (+-0.72) &        0.01 (+-0.02) &         2.77 (+-0.17) &                5.66 (+-0.22) &    3.37  (+-0.13)  & 0.9894 (+-0.0008) \\

&    & \textbf{AlexNet\_2048\textsubscript{reduced}}  & \textbf{SVM (RBF)} &  \textbf{6.75 (+-0.42)} &        \textbf{0.00 (+-0.00)} &         \textbf{3.46 (+-0.12)} &                 \textbf{4.85  (+-0.11)} &   \textbf{2.74 (+-0.18)} & \textbf{0.9913 (+-0.0009)} \\

GPDS-300 & 5 & AlexNet\_2048 & Linear SVM & 31.48 (+-1.32) &        0.00  (+-0.00) &         1.87 (+-0.15) &                 9.39 (+-0.44) & 5.41 (+-0.40)  &    0.9760  (+-0.0017) \\

&   & AlexNet\_2048\textsubscript{reduced}  &SVM (RBF) & 26.33 (+-0.99) &        0.00 (+-0.00) &         1.74 (+-0.11) &                 8.04  (+-0.20) & 4.53 (+-0.14) &   0.9817 (+-0.0008)  \\

&   14 & AlexNet\_2048 &Linear SVM & 10.42 (+-0.58) &        0.00 (+-0.01) &         4.73  (+-0.19) &                 7.00 (+-0.15) &   4.17 (+-0.25) & 0.9823 (+-0.0012)  \\

&   & \textbf{AlexNet\_2048\textsubscript{reduced}}  &\textbf{SVM (RBF)} & \textbf{6.55 (+-0.25)} &        \textbf{0.00  (+-0.01)} &         \textbf{5.13 (+-0.20)} &                 \textbf{5.75 (+-0.19)}&   \textbf{3.47 (+-0.16)} & \textbf{0.9871 (+-0.0007)}\\

\bottomrule
\end{tabular}
}
\end{table*}

\begin{table}
\centering
\caption{Comparison with state-of-the art on the GPDS dataset (errors and standard deviations in \%)}
\label{table:soa_gpds}
\resizebox{\columnwidth}{!}{%
\begin{tabular}{llllr}
\toprule
 Reference & Dataset& \begin{tabular}[x]{@{}c@{}}\#samples\\per user\end{tabular}  &Features \& (Classifier)  &  EER\\
\midrule
Vargas et al \cite{vargas_off-line_2010}  &GPDS-100 & 5 &Wavelets (SVM) & 14.22 \\
Vargas et al \cite{vargas_off-line_2011} & GPDS-100 &10 &LBP, GLCM (SVM)&  9.02 \\

Hu and Chen \cite{hu_offline_2013}& GPDS-150 &10 &LBP, GLCM, HOG (Adaboost) & 7.66\\
Yilmaz \cite{yilmaz_offline_2015}  & GPDS-160 &12&LBP (SVM) & 9.64\\
Yilmaz \cite{yilmaz_offline_2015}  &GPDS-160 &12&LBP, HOG  (Ensemble of SVMs)& 6.97\\
Hafemann et al \cite{hafemann_ijcnn_2016} & GPDS-160 & 14 & WI-learned with a CNN (SVM) &  10.70 \\

\midrule

\textbf{Present work} &\textbf{GPDS-160} & \textbf{5}  &\textbf{WI-learned with a CNN  (SVM)}&  \textbf{3.83 (+- 0.33)}\\
\textbf{Present work} &\textbf{GPDS-160} & \textbf{14}  &\textbf{WI-learned with a CNN  (SVM)}& \textbf{2.74 (+- 0.18)}\\

\textbf{Present work} &\textbf{GPDS-300} & \textbf{5}  &\textbf{WI-learned with a CNN  (SVM)}&  \textbf{4.53 (+- 0.14)}\\
\textbf{Present work} &\textbf{GPDS-300}  & \textbf{14}  &\textbf{WI-learned with a CNN  (SVM)}&  \textbf{3.47 (+- 0.16)}\\

\bottomrule
\end{tabular}
}
\end{table}

\begin{figure}
\centering
\includegraphics[width=\columnwidth]{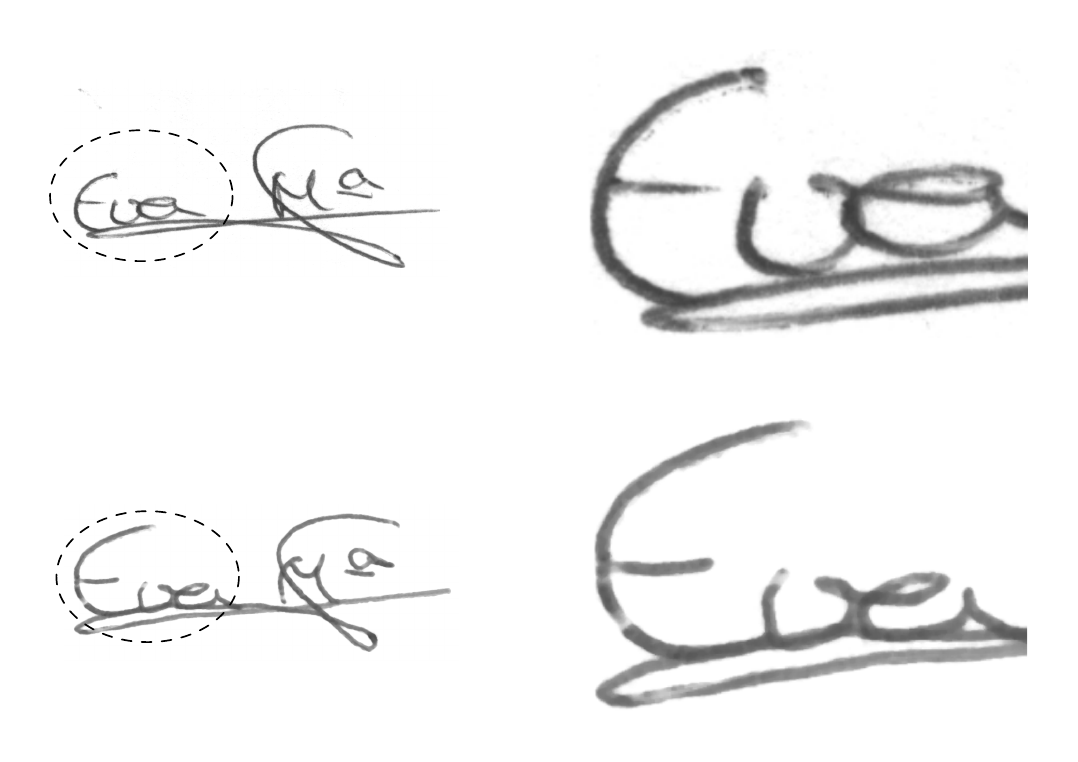}
\caption{Details of a genuine signature (top), and a skilled forgery (bottom) that are mapped close in the embedding space $\phi(.)$}
\label{fig:closeup}
\end{figure}

The zoomed-in area in figure \ref{fig:tsne_genuine_fogery} shows skilled forgeries made for a particular user that were represented close to genuine signatures from this user. We can see that these forgeries have a close resemblance to the genuine signatures of the user in its overall shape. This is further examined in figure \ref{fig:closeup}, where we take a genuine signature and a skilled forgery from this user, that were close in the embedding space. We can notice that the overall shape of the skilled forgery is similar to the genuine signature, but looking in the details of the strokes we can see that the line quality of the skilled forgery is much worse. We have noticed the same behavior with other users / signatures. This suggests that the features learned by the network are useful in distinguishing the signatures in an ``overall shape'', but do not capture important properties of forgeries, such as the quality of the pen strokes, and therefore are particularly not discriminative to slowly-traced forgeries.

Table \ref{table:results} shows the detailed result of the experiments in the exploitation set. We noticed that the performance was very good in terms of equal error rates, but that the global decision thresholds from the validation set are only effective in a few situations - in particular, when using the same number of samples as used in the validation set, where the threshold was selected.  We also notice that the performance using a global threshold is significantly worse than using per-user thresholds. Note that the values of EER refer to the scenario where the decision threshold is optimally selected, which in itself is a hard task (given the low number of samples per user), and which we did not explore in this paper. Finally, table \ref{table:soa_gpds} compares our best results (using EER\textsubscript{user-thresholds}) with the state-of-the-art. We noticed a big drop in classification errors (as measured by Equal Error Rate) compared to the literature, even when using only 5 samples per user for the Writer-Dependent training.

\section{Conclusions}

In this work, we presented a detailed analysis of different architectures for learning representations for offline signature verification. We showed that features learned in a writer-independent format can be very effective for signature verification, achieving a large improvement of state-of-the-art performance in the GPDS-160 dataset, with 2.74\% Equal Error Rate compared to 6.97\% reported in the literature. We also showed that writer-dependent classifiers trained with these features can perform very well even with limited number of samples per user (e.g. 5 samples) and linear classifiers.

Our analysis of the signature representations showed that the learned features are mostly useful to distinguish signatures on a ``general appearance'' instead of finer details. This make these features relevant for distinguishing random forgeries and skilled forgeries that are made with quick motion (but do not perfectly capture the overall aspect of the genuine signature). On the other hand, it makes these features less discriminant to slowly-traced skilled forgeries, where the forgery looks like a genuine signature in the overall shape, but has a poor line quality. Future work can investigate the combination of such features with features particularly targeted to discriminate the quality of pen strokes.

\section*{Acknowledgment}

This research has been supported by the CNPq grant \#206318/2014-6.

\bibliographystyle{IEEEtran}
\bibliography{biblio}

\end{document}